\RequirePackage{fix-cm}
\documentclass[twocolumn]{svjour3}          % twocolumn
\smartqed  % flush right qed marks, e.g. at end of proof

\usepackage{cite}
\usepackage[pdftex]{graphicx}
\usepackage{ragged2e}
\usepackage[tight,footnotesize]{subfigure}
\usepackage{rotating}
\usepackage{graphicx}
\usepackage{amsmath,amssymb} % define this before the line numbering.
\usepackage{array}
\usepackage{multirow}
\usepackage{colortbl}
\usepackage{bm}
\usepackage{amsfonts}
\usepackage{pifont}
\usepackage{xspace}
\usepackage{etoolbox}
\usepackage{overpic}
\usepackage{booktabs}
\usepackage{color}
\usepackage{microtype}
\usepackage{float}
\usepackage{multirow}
\usepackage{graphicx}
\usepackage{array}
\usepackage{makecell}
\usepackage{url}

\newcommand{\etal}{\textit{et al}.}

\newcommand{\orcid}[1]{\href{https://orcid.org/#1}{\includegraphics[width=10pt]{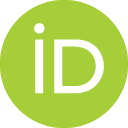}}}

\def\etal{{\em et al}}

\usepackage{hyperref}
\hypersetup{breaklinks=true,citecolor=blue, colorlinks}

\DeclareGraphicsExtensions{.pdf,.jpg,.png}

\usepackage{silence}
\journalname{Research Article}

\begin{document}

\title{Temporal Consistency-Aware Text-to-Motion Generation}

\titlerunning{Short form of title}        % For running head

\author{Hongsong Wang\textsuperscript{1,2}        \and
	Wenjing Yan\textsuperscript{1,2} \and 
	Qiuxia Lai\textsuperscript{3*} \and
	Xin Geng\textsuperscript{1,2}
}
%\author{Hongsong Wang\textsuperscript{1,2} \orcid{0000-0002-9464-1778}        \and
	%  Wenjing Yan\textsuperscript{1,2} \orcid{0000-0000-0000-0000} \and 
	%  Qiuxia Lai\textsuperscript{3} \orcid{0000-0001-6872-5540} \and
	%  Xin Geng\textsuperscript{1,2} \orcid{0000-0001-7729-0622}
	%}

\authorrunning{F. Author \etal} % if too long for running head

% \institute{
	% Hongsong Wang, Wenjing Yan and Xin Geng are with School of Computer Science and Engineering, Southeast University, Nanjing 210096, China, Key Laboratory of New Generation Artificial Intelligence Technology and Its Interdisciplinary Applications (Southeast University), Ministry of Education, China. 
	% (Email: hongsongwang@seu.edu.cn, yanwl2119@seu.edu.cn, xgeng@seu.edu.cn). \\
	% Qiuxia Lai is with State Key Laboratory of Media Convergence and Communication, Communication University of China, Beijing 100024, China. 
	% (Email: qxlai@cuc.edu.cn). \\
	% Corresponding author: Qiuxia Lai. \\
	% The code is available at \url{https://github.com/Giat995/Temporal-Consistency-Aware-Text-to-Motion-Generation/}
	% }
\institute{
	\textsuperscript{1} School of Computer Science and Engineering, Southeast University, Nanjing, China.\\
	\textsuperscript{2} Key Laboratory of New Generation Artificial Intelligence Technology and Its Interdisciplinary Applications (Southeast University), Ministry of Education, China.\\
	\textsuperscript{3} State Key Laboratory of Media Convergence and Communication, Communication University of China, Beijing, China.\\
	\textsuperscript{*} \textit{Corresponding author:} Qiuxia Lai, Email: \texttt{qxlai@cuc.edu.cn}
}

\date{Received: date / Accepted: date}

\maketitle

\begin{abstract}
	Text-to-Motion (T2M) generation aims to synthesize realistic human motion sequences from natural language descriptions. While two-stage frameworks leveraging discrete motion representations have advanced T2M research, they often neglect cross-sequence temporal consistency, i.e., the shared temporal structures present across different instances of the same action. This leads to semantic misalignments and physically implausible motions. To address this limitation, we propose TCA-T2M, a framework for temporal consistency-aware T2M generation. Our approach introduces a temporal consistency-aware spatial VQ-VAE (TCaS-VQ-VAE) for cross-sequence temporal alignment, coupled with a masked motion transformer for text-conditioned motion generation. Additionally, a kinematic constraint block mitigates discretization artifacts to ensure physical plausibility. Experiments on HumanML3D and KIT-ML benchmarks demonstrate that TCA-T2M achieves state-of-the-art performance, highlighting the importance of temporal consistency in robust and coherent T2M generation. 
	
	% Please provide 4 to 6 keywords which can be used for indexing purposes.
	
\end{abstract}

\section{Introduction}
\label{sec:intro}

Text-to-Motion (T2M) aims to translate natural language descriptions into realistic human motion sequences, with applications that span virtual reality~\cite{guo2022generating}, film production~\cite{kappel2021high}, and human-robot interaction~\cite{antakli2018intelligent,koppula2013learning,koppula2015anticipating}. Recent advances in T2M research have witnessed a surge of research interest, where methods can be broadly categorized into end-to-end frameworks and two-stage architectures. While early end-to-end approaches explore direct feature alignment between language and motion spaces~\cite{ahuja2019language2pose, ghosh2021synthesis}, two-stage architectures, which decomposes the task into motion representation learning and conditional motion generation~\cite{guo2022generating, guo2024momask, lin2018generating, pinyoanuntapong2024mmm}, have emerged as the dominant paradigm due to their superior performance in semantic fidelity and motion realism. In this setup, a text-independent motion representation space is first learned by capturing the intrinsic characteristics of human motion, followed by text-conditioned generation that bridges the semantic gap, producing coherent motion sequences aligned with language descriptions.

Two-stage T2M models can be classified into three main paradigms based on their types of motion representation. 
AE-based methods~\cite{lin2018generating} learn compact motion representations through deterministic encoding. Although their outputs can serve as faithful reconstructions of input actions, they inherently suffer from limited diversity in motion generation. 
VAE-based methods~\cite{athanasiou2022teach,guo2022generating,lin2023being,petrovich2022temos} introduce probabilistic latent spaces to enable diverse motion synthesis through stochastic sampling. However, their continuous latent spaces pose challenges for text-to-motion alignment due to the nonlinear and high-dimensional nature of motion dynamics. This often leads to unreliable mappings in encoder-decoder architectures~\cite{guo2022tm2t}. 
VQ-VAE-based methods~\cite{guo2022tm2t,guo2024momask,hosseyni2025bad,liu2024emage,liu2024towards,pinyoanuntapong2024bamm,pinyoanuntapong2024mmm,yang2024unimumo,yi2023generating,zhang2023generating} tackle this by discretizing continuous motion features into codebook tokens, which simplifies the generation process while enhancing computational efficiency. However, the discretization process disrupts the temporal continuity of motion, which often introduces semantic discontinuities between discrete tokens that manifest themselves as unnatural gait artifacts such as ``leg sliding''. 
Moreover, while existing T2M models focus on learning motion representations at the instance level, they overlook a fundamental property of human motion, i.e., shared temporal structures across different realizations of the same action. This cross-sequence temporal consistency is crucial for robust motion generation but remains largely unaddressed in current frameworks.
% More critically, existing paradigms rely solely on per-sample reconstruction loss, ignoring temporal correspondences across motion sequences. As a result, they fail to capture temporal dynamics shared across instances of the same action, leading to inconsistent motion patterns within action categories.

\begin{figure*}[t]
	\centering
	\includegraphics[width=1.0\linewidth]{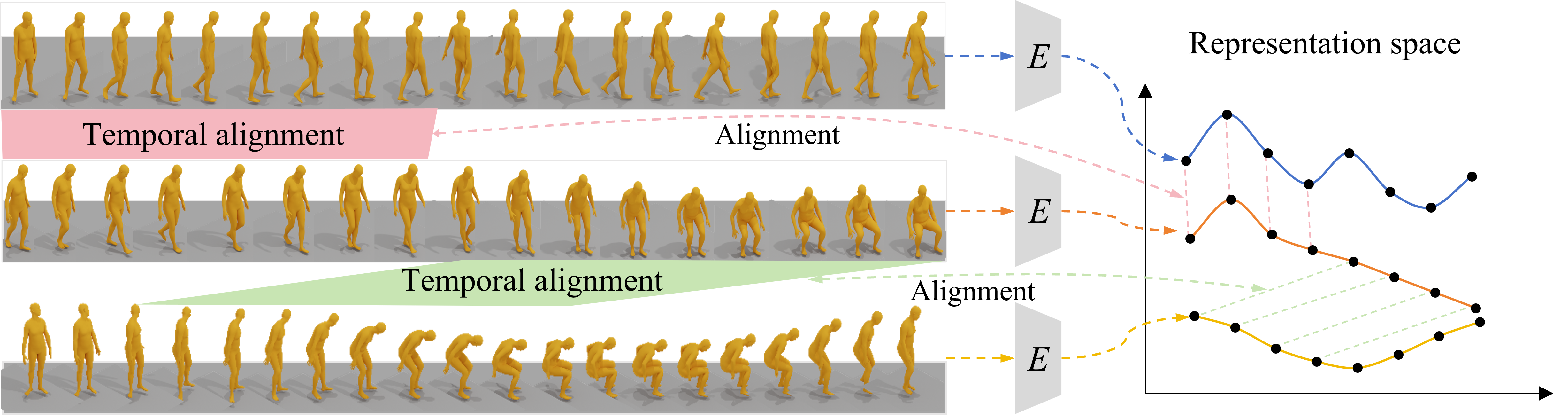}
	\caption{Illustration of temporal consistency across three distinct human action sequences. (a) A person walks forward; (b) A person walks and sits down; (c) A person sits down and stands up. Despite differences in kinematic details, these sequences exhibit shared temporal structures. Enforcing temporal alignment in the latent space ensures that motion representation encoder $E$ maps corresponding action phases across sequences to similar representations. This constraint enables the learned motion representation to capture semantic information while preserving temporal consistency, which is essential for the subsequent text-conditioned motion generation stage in T2M. 
		% See Section \ref{sec:intro} for details.
		% The same actions across different sequences are temporally aligned, and their feature representations projected into the latent space via encoder E maintain alignment. Different segments of the same action map to similar positions in the representation space, indicating that the encoder effectively captures semantic information while preserving temporal consistency in the low-dimensional space. Such representation is crucial for tasks like action recognition, classification, and generation.
	}
	\label{fig:cycle_consistency}
\end{figure*}

% This fundamental issue, \ie, cross-sequence temporal inconsistency, undermines current motion generation paradigms. 
% A fundamental limitation of motion representation learning in current T2M paradigms lies in cross-sequence temporal inconsistency.
Human motion is inherently a spatiotemporally continuous physical process, where semantic consistency is not confined to individual sequences but emerges across multiple realizations of the same action. As illustrated in Fig.~\ref{fig:cycle_consistency}, diverse motion sequences, such as forward walking, walk-to-sit, and sit-to-stand, exhibit shared temporal structures despite differences in kinematic details. These shared patterns include temporal landmarks such as the timing of foot contacts or body weight shifts, which form the semantic core of an action. Effective motion encoders are expected to capture these cross-sequence temporal correspondences to align sequences according to their underlying motion semantics. This alignment enables the latent space to reflect the fundamental invariants of human motion, rather than the idiosyncrasies of individual instances. However, existing methods typically neglect this requirement and treat each motion sequence in isolation,  which fails to enforce temporal consistency across related actions. 
Moreover, neglecting temporal consistency can also introduce artifacts that undermine physical plausibility.

Motivated by these insights, we propose TCA-T2M, a framework for temporal consistency-aware T2M generation that integrates cyclic temporal alignment constraints into discrete representation learning. Our approach contains two major components: the temporal consistency-aware spatial VQ-VAE (TCaS-VQ-VAE) integrated with the kinematic constraint block, and the masked motion transformer.
TCaS-VQ-VAE introduces cross-sequence temporal alignment constraints at the decoder input through contrastive learning in the latent space, which encourages the encoder to capture and align temporal consistency features across different instances of the same action. This design promotes both discriminative token learning for modeling inter-sequence variations and temporal coherence within each sequence, enabling the model to distinguish fine-grained action differences. 
To further mitigate gait artifacts caused by representation discretization, inspired by Ref.~\cite{li2024lodge}, we introduce a kinematic constraint block into TCaS-VQ-VAE, which enforces kinematic consistency by applying physics-based motion constraints to reduce foot-sliding and enhance the physical realism of generated motions. 
Finally, the masked motion transformer adopts the design of Ref.~\cite{guo2024momask} for its proven effectiveness, which employs dynamic masking and text-conditioned iterative refinement for phase-wise motion generation. 
Extensive experiments on the HumanML3D~\cite{guo2022generating} and KIT-ML~\cite{plappert2016kit} datasets demonstrate that our approach achieves state-of-the-art performance across both quantitative and qualitative benchmarks, significantly improving motion fidelity and temporal coherence.
% we investigate temporal-consistent motion representation learning for preserving cross-temporal coherence in motion sequences. We propose TCA-T2M, a novel framework integrating cyclic temporal alignment constraints with discrete representation learning. Our architecture comprises two core components: (1) the Temporal Consistency-aware Spatial VQ-VAE introduces temporal alignment constraints at the decoder input, compelling the encoder to learn cross-sequence temporal consistency features through contrastive latent space representation learning of different sequences within the same action category. Notably, the incorporation of TCC endows motion tokens with discriminative consistency—a critical property where TCC-driven contrastive learning enhances inter-sequence feature separability while preserving intra-sequence temporal coherence. This dual capability enables learned tokens to explicitly differentiate motion patterns across sequences while maintaining logical continuity within individual motions. The refined token discriminability fundamentally empowers the second component: the Masked Motion Transformer. This module employs dynamic mask generation strategies and text-conditioned iterative optimization for phase-wise motion sequence prediction. To mitigate gait discontinuities induced by VQ-VAE discretization artifacts, we introduce a Foot Refinement Block within TCaS VQ-VAE. This block enforces kinematic consistency through physics-based motion constraints, effectively suppressing foot-sliding artifacts while preserving motion realism.
In summary, the main contributions are as follows:

1) We propose TCA-T2M, a framework for temporal consistency-aware T2M generation that integrates cyclic temporal alignment constraints into discrete representation learning.

2) We introduce temporal consistency-aware spatial VQ-VAE, a pioneering effort to embed cyclic temporal alignment constraints into motion token learning, which enables cross-sequence temporal alignment across diverse motion instances within the same action category. 
% \item We develop a Kinematic Constraint Block that enforces joint acceleration continuity, ensuring physically plausible motion outputs. %  and mitigating discretization-induced artifacts

% Our principal contributions are threefold: First, we pioneer the integration of temporal consistency features into T2M feature extraction by proposing the Temporal Consistency-aware Spatial VQ-VAE. This architecture overcomes the local encoding limitations of conventional VQ-VAE through latent space representation learning that enforces cross-sequence temporal pattern commonalities across different motion instances within the same category. Second, to resolve the physical implausibility in VQ-VAE-generated motions, we develop a Kinematic Constraint Block—a refinement network incorporating joint acceleration continuity penalties. Third, extensive empirical validation demonstrates state-of-the-art performance on both qualitative and quantitative benchmarks using HumanML3D~\cite{guo2022generating} and KIT-ML~\cite{plappert2016kit} datasets.

\section{Related Work}
\label{related_work}

\textit{Text-to-Motion generation} Text-to-motion (T2M) generation aims to synthesize natural human motions from textual descriptions. The core challenge lies in bridging the gap between the discrete nature of language and the continuous, spatiotemporal complexity of human motion~\cite{weng2025realign,tan2024sopo}.
% establishing cross-modal mappings between textual semantics and human motion sequences, rooted in the inherent contradiction between the spatiotemporal complexity of motions and the discreteness of textual representations. 
Existing methods can be broadly categorized into single-stage and two-stage approaches. 

Early single-stage methods directly map text to motion in an end-to-end manner. For instance, Language2Pose~\cite{ahuja2019language2pose} aligns action and language features in a shared embedding space, while Ghosh et al~\cite{ghosh2021synthesis} designed a hierarchical two-stream sequential model to capture finer joint-level dependencies. 
While these methods are conceptually simple and computationally efficient, they often struggle to generate high-fidelity, temporally coherent motions.

Two-stage approaches decompose the problem into learning a motion representation space and generating motion sequences conditioned on text. The development of motion representations has followed distinct lines of research.
Early work, such as Seq2Seq~\cite{lin2018generating}, uses deterministic autoencoders to embed motion sequences into latent spaces. However, fixed mappings often limit the diversity of generated motions. Variational methods~\cite{guo2022generating, petrovich2022temos} introduce stochasticity via VAEs, enhancing diversity but facing challenges such as non-convex objectives, optimization instability, and the lack of explicit temporal constraints, which often lead to fragmented temporal coherence in generated motions. Moreover, continuous latent spaces require high-dimensional representations to capture fine-grained details, yet existing methods struggle to balance generation quality with latent space dimensionality due to computational constraints.
More recently, discrete token-based approaches, such as VQ-VAE models~\cite{van2017neural, pinyoanuntapong2024mmm}, have been adopted for motion representation learning. By discretizing motion sequences into tokens, these approaches reduce latent dimensionality and exploit local token correlations for efficient learning. However, the discretization process can disrupt the physical continuity of motion, leading to semantic discontinuities between adjacent tokens and introducing artifacts such as unnatural gait transitions. 
Moreover, while current T2M models focus on instance-level motion representations, they often overlook an important property of human motion, i.e., shared temporal structures across different instances of the same action. This cross-sequence temporal consistency is vital for generating realistic, coherent motions but remains underexplored in existing frameworks. 

\noindent\textit{Cycle-Consistent learning} 
Cycle-consistent learning offers a natural solution to cross-sequence temporal alignment by enforcing structural consistency through cyclic mappings. It has proven effective in various computer vision tasks, such as image matching~\cite{zhou2015flowweb, zhou2016learning, zhou2015multi}, segmentation~\cite{wang2013image, wang2014unsupervised}, video generation~\cite{wang2018video}, and video representation learning~\cite{dwibedi2019temporal}, where preserving structural coherence across domains is essential. For example, FlowWeb~\cite{zhou2015flowweb} leverages cycle-consistent flow fields to optimize dense correspondences across image collections, while CycleGAN~\cite{zhu2017unpaired} applies cycle consistency with adversarial training for image style transfer, effectively suppressing artifacts such as jitter and foot sliding. 
TCC~\cite{dwibedi2019temporal} learns self-supervised video frame embeddings using a temporal cycle-consistency loss, enabling temporal alignment between videos by matching nearest-neighbor frames in the learned embedding space.
Beyond vision tasks, cycle-consistency has been applied in cross-task learning, such as visual question answering~\cite{li2018visual, shah2019cycle, tang2017question} and vision-language navigation~\cite{wang2022counterfactual}, improving semantic alignment through joint bidirectional training.

Inspired by these successes, we extend cycle-consistent learning to the domain of text-to-motion generation. We propose a temporal consistency constraint that enforces cross-sequence alignment of motion representations within the same action class, addressing a fundamental limitation of prior T2M methods. By regularizing latent spaces to respect shared temporal structures, our approach enables the generation of semantically coherent and physically plausible motion sequences.

\noindent\textit{Temporal consistency in video synthesis} 
Temporal consistency is a crucial issue in video synthesis, as even minor inconsistencies can severely degrade perceptual quality. Chu et al.~\cite{chu2020learning} introduced a self-supervised temporal constraint that prevents long-term drift and flickering in video generation. CCVS~\cite{le2021ccvs} combines the autoregressive framework with a learnable optical flow module to enforce temporal consistency. TECO~\cite{yan2023temporally} learns compressed representations to enable efficient video generation with long-range temporal consistency. MagicAnimate~\cite{xu2024magicanimate} introduces an appearance encoder to enhance temporal consistency across video frames. FlowVid~\cite{liang2024flowvid} proposes a consistent video-to-video synthesis framework by leveraging temporal optical flow clues. FETV~\cite{liu2023fetv} proposes fine-grained metrics to measure temporal consistency, motion accuracy, and semantic alignment. Although improving temporal consistency in video synthesis has been studied, there is still limited work addressing temporal consistency in motion synthesis, and our work fills this gap.

\section{Method}
\label{sec:method}
\begin{figure*}[t]
	\centering
	\includegraphics[width=1.0\linewidth]{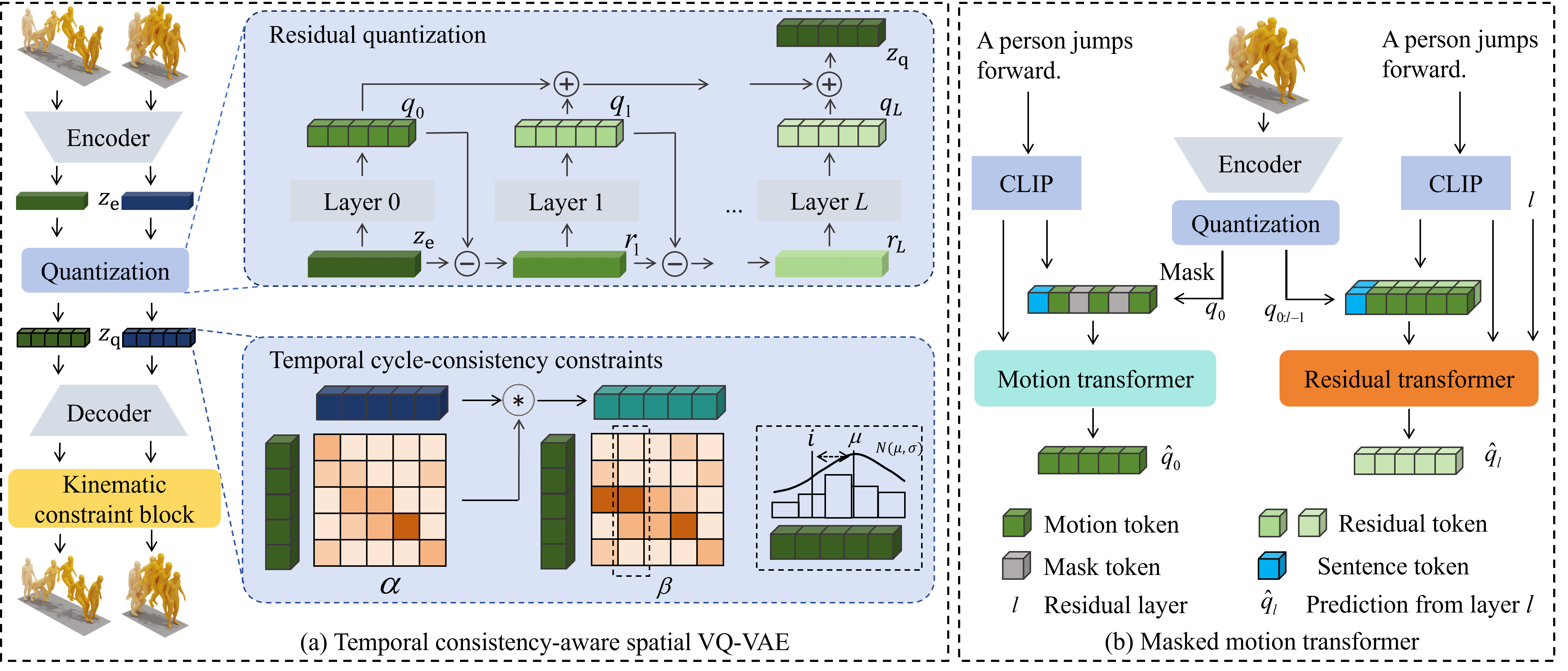}
	\caption{Method overview. (a) Temporal consistency-aware spatial VQ-VAE employs hierarchical residual quantization to discretize motion features, incorporates cycle-consistency constraints to enforce temporal coherence, and utilizes a kinematic constraint block to refine motion details. (b) Masked motion transformer adopts a dual-transformer structure for cross-modal text-motion synthesis. Specifically, the motion transformer restores masked motion tokens under CLIP text guidance, while the residual transformer predicts successive residual tokens using textual context and preceding token sequences. }
	\label{fig:overview}
\end{figure*}

\subsection{Overview}
\label{overview}

Our objective is to develop a T2M framework capable of generating high-fidelity 3D human motion sequences from textual descriptions. Given a text prompt $c$, the system produces a motion sequence $m_{1:N}$ of length $N$, where $m_i \in \mathbb{R}^D$, and $D$ represents the dimension of the motion feature. 
We propose temporal consistency-aware text-to-motion (TCA-T2M), a unified architecture that consists of the temporal consistency-aware spatial VQ-VAE (TCaS-VQ-VAE) and a dynamic mask-guided generation framework (c.f.~Fig.~\ref{fig:overview}). Our approach addresses two key limitations of existing discrete representation approaches, namely temporal misalignment and kinematic implausibility, through cross-sequence temporal constraint modeling and physics-aware optimization. The architecture consists of two core components: (1) TCaS-VQ-VAE, a VQ-VAE-based model that learns cross-sequence temporal patterns via cycle-consistency constraints, incorporates residual quantization to minimize information loss, and incorporates a kinematic constraint block to enforce joint acceleration continuity constraints to enhance motion physicality; and (2) masked motion transformer architecture, a two-stage transformer architecture that leverages dynamic mask unrolling strategies to achieve hierarchical text-motion alignment. 
%(2) \textbf{Kinematic Constraint Block} that enforces joint acceleration continuity constraints to enhance motion physicality;

The remainder of this section is organized as follows. Section~\ref{sec:tcas_vq_vae} details the temporal consistency-aware motion representation learning, residual quantization, and the kinematic constraint block in TCaS-VQ-VAE. Section~\ref{masked motion transformer} presents the mask-guided generation framework of the masked motion transformer and also outlines the inference pipeline.
% Finally, Section~\ref{sec:inference} outlines the inference pipeline.

\subsection{Temporal Consistency-aware Spatial VQ-VAE}
\label{sec:tcas_vq_vae}
The temporal consistency-aware spatial VQ-VAE (TCaS-VQ-VAE) aims to construct a discrete representation space that captures cross-sequence temporal invariances while enhancing motion reconstruction quality through cycle-consistency regularization and residual quantization. Conventional VQ-VAE architectures process input motion sequences $\bm{m}_{1:N} \in \mathbb{R}^{N \times D}$ through stacked convolutional layers to extract spatiotemporal features, yielding continuous latent representations $\bm{z}_{\mathrm{e}} \in \mathbb{R}^{n \times d}$, where $n/N$ is the downsampling ratio, and $d$ is the feature dimension of the latent vectors. These representations undergo quantization via nearest-neighbor search in a learnable codebook $\bm{C} = \{e_1, \ldots, e_K\} \in \mathbb{R}^{K \times d}$, producing discrete token sequences $\bm{z}_{\mathrm{q}} = Q(\bm{z}_{\mathrm{e}}) \in \mathbb{R}^{n \times d}$. The decoder reconstructs motion sequences through $\bm{\hat{m}} = D(\bm{z}_{\mathrm{q}})$. This method minimizes the reconstruction error through a reconstruction loss, constrains the distance between the encoder output and the codebook via a quantization loss, and uses a commitment loss to encourage the codebook vectors to converge towards the encoder's output distribution. This loss is defined as
\begin{equation}\label{VQ loss}
	\mathit{L}_{\mathrm{vq}}=\Vert \bm{m}-\bm{\hat{m}}\Vert^2_2+\Vert \text{sg}[\bm{z}_{\mathrm{e}}]-\bm{z_q}\Vert^2_2+\gamma\Vert \bm{z}_{\mathrm{e}}-\text{sg}[\bm{z}_{\mathrm{q}}]\Vert^2_2, 
\end{equation}
where $\text{sg}$ denotes the stop-gradient operation, and $\gamma$ is the hyperparameter for the commitment loss.

\noindent\textit{Temporal Cycle-Consistency Constraint}
To enable VQ-VAE to learn cross-sequence temporal consistency during motion reconstruction, this section introduces cycle-consistency constraints. By establishing reversible mapping relationships between motion sequences, the encoder is compelled to learn global temporal pattern consistency across different sequences. Specifically, motion sequences are categorized by action labels derived from textual descriptions. Consider two distinct motion sequences $\bm{U}_{1:T}$ and $\bm{V}_{1:T}$ from the same action category. Here, the subscript $T$ denotes the temporal dimension, where $T$ represents the total number of frames or time steps in the sequences. After feature extraction and quantization, we obtain $\bm{U}_\mathrm{q} = Q(E(\bm{U})) = \{u_1, u_2, \ldots, u_n\} $ and $ \bm{V}_\mathrm{q} = Q(E(\bm{V})) = \{v_1, v_2, \ldots, v_n\}$. Here, q, $Q$, and $E$ represent quantization, quantizer, and encoder respectively. To verify the cycle consistency of a point $u_i \in U$, we first identify its nearest neighbor in sequence $\bm{V}$ as $v_j = \arg\min_{v \in V} \Vert u_i - v \Vert $, and then determine the nearest neighbor of $v_j$ in sequence $\bm{U}$ as $u_k = \arg\min_{u \in U} \Vert v_j - u \Vert $. If $i=k$, $u_i$ is classified as cycle-consistent, indicating bidirectional one-to-one mapping capability between sequences. By maximizing the ratio of such cycle-consistent points, i.e., the proportion of points establishing mutual mappings between sequences of the same action category—the model captures cross-sequence temporal alignment patterns, thereby addressing the single-sequence modeling limitation inherent in conventional VQ-VAE.

Cycle-consistency computation inherently involves discrete nearest-neighbor matching, which is not differentiable and precludes end-to-end optimization. To address this, we model cycle-consistency through dual perspectives of classification and regression. For each point $u_i$ in sequence $\bm{U}_\mathrm{q}$, we compute its similarity to all points $v_j$ in sequence $\bm{V}_\mathrm{q}$ and derive the similarity distribution $\alpha$ via softmax transformation
\begin{equation}\label{similarity_cls}
	\begin{split}
		\alpha_j=\frac{\mathrm{e}^{s_j}}{\sum ^n_{k=1}\mathrm{e}^{s_k}},
	\end{split}
\end{equation}
$\text{where}\;s_j=-\Vert v_j-u_i\Vert^2.$ The temporally aligned point $\tilde{v}$ in sequence $\bm{V}$ is obtained as $\tilde{v} = \sum_{k} \alpha_k v_k$. By treating each point in $\bm{U}_\mathrm{q}$ as an independent class, cycle-consistency detection is reformulated as a nearest-neighbor classification task. We construct an $n$-class classification framework to predict the cyclic index of $u_i$ in $\bm{V}_\mathrm{q}$ and its reciprocal mapping back to $\bm{U}_\mathrm{q}$, formalizing the cycle classification loss as
\begin{equation}\label{classification}
	\mathit{L}_\mathrm{cls} = -\sum_{k=1}^n \mathbb{I}(k=i) \log(\alpha_j),
\end{equation}
% where $\mathbb{I}(\cdot)$ is the indicator function, equal to 1 when $k=i$.
where $\mathbb{I}(\cdot)$ is the indicator function, and its value is 1 when $k=i$.

While cycle classification loss verifies the temporal mapping's closed-loop property through probabilistic classification, it exclusively focuses on endpoint category matching while neglecting spatial offsets of mapped points along the temporal axis. Specifically, the cycle classification loss employs cross-entropy to constrain whether the index $k$ of the soft nearest neighbor $\tilde{v}$ in $\bm{V}_\mathrm{q}$ matches the original index $i$ when reciprocally mapped to $\bm{U}_\mathrm{q}$. However, this approach fails to explicitly optimize the temporal alignment distance between $\tilde{v}$'s projected position on $\bm{U}_\mathrm{q}$'s time axis and $i$, potentially leading to significant temporal misalignment errors. To address this, we introduce a cycle regression loss that directly optimizes temporal alignment accuracy via a dual-constraint mechanism. First, we compute the similarity $\beta$ between the mapped point $\tilde{v}$ and points in $\bm{U}_\mathrm{q}$
\begin{equation}\label{similarity_reg}
	\begin{split}
		\beta_j=\frac{\mathrm{e}^{s_j}}{\sum ^n_{k=1}\mathrm{e}^{s_k}},
	\end{split}
\end{equation}
where$\;s_j=-\Vert \widetilde{v}-u_j\Vert^2.$ The distribution $\beta$ characterizes the soft alignment region of $\tilde{v}$ on the time axis of $\bm{U}_\mathrm{q}$. Ideally, $\beta$ should exhibit a peak around $i$, indicating temporal alignment between $\tilde{v}$ and $u_i$. To enforce this, we apply a Gaussian prior on $\beta$ by minimizing the distance between the predicted position $\mu = \sum_{j} \beta_j \cdot j$ and the ground true index $i$, while regularizing the variance $\sigma^2 = \sum_{j} \beta_j \cdot (j - \mu)^2$ to control the shape of the distribution shape and suppress overfitting to local noise. The cycle regression loss is computed using the mean squared error (MSE), which is mathematically expressed as
\begin{equation}\label{regression}
	\mathit{L}_\mathrm{reg\_mse}=\frac{\lvert i-\mu \rvert^2}{\sigma^2}+\lambda \log(\sigma),
\end{equation}
where $\lambda$, $i$ and $\mu$ denote the regularization weight, mean and variance, respectively. This dual-constraint mechanism ensures that the peak of $\beta$ is strictly localized around $i$ with controlled diffusion along the temporal axis, achieving both local mapping precision and noise robustness. In addition to the MSE formulation, the cycle regression loss can also be implemented using the Huber loss to enhance robustness against outliers:
\begin{equation}
	\mathit{L}_{\mathrm{reg\_huber}} = \frac{1}{\sigma^2} \cdot L_{\delta}(i-\mu) + \lambda \log(\sigma),
\end{equation}
where $\mathit{L}_{\delta}$ is the Huber loss, which is defined as
\begin{equation}
	\mathit{L}_{\delta} = \left\{
	\begin{array}{lc}
		\frac{1}{2} (i-\mu)^2, & \text{if}~~\vert i-\mu \vert \leq  \delta\\
		\delta(\vert i-\mu \vert - \frac{1}{2} \delta), & \text{otherwise}.
	\end{array}
	\right.
\end{equation}
where $\delta$ is a hyperparameter that controls the behavior of $\mathit{L}_{\delta}$. Based on the ablation results in Section \ref{sec:ablation_study} and Table~\ref{tab:ablation}, we select the MSE-based cycle regression loss $\mathit{L}_{\mathrm{reg\_mse}}$ as the final formulation for temporal cycle-consistency constraint loss $L_\mathrm{tcc}$.

% Combining both losses, the total temporal cycle-consistency loss becomes:
% \begin{equation}\label{tcc_loss}
	%     \mathcal{L}_{tcc}=\lambda_c\mathcal{L}_{cls}+\lambda_r\mathcal{L}_{reg},
	% \end{equation}
% where $\lambda_c$ and $\lambda_r$ are hyperparameters for the cycle classification and regression losses, respectively.

\noindent\textit{Residual Quantization}
To mitigate information loss caused by quantization in traditional VQ-VAE models, which hampers motion reconstruction quality, we introduce a residual quantization mechanism inspired by Ref.~\cite{guo2024momask}. This method enhances detail preservation in discrete representations through layered error compensation. Unlike standard VQ-VAE, which performs a single-stage quantization on continuous latent features, our method decomposes the process into multi-stage residual approximations. Specifically. the encoder's continuous output $\bm{z}_\mathrm{e}$ is first quantized by the $0$-th quantizer to generate a coarse code $\bm{q}_0$ and a residual $\bm{r}_1 = \bm{z}_\mathrm{e} - \bm{q}_0$. Each subsequent layer quantizes the remaining residuals iteratively: $\bm{q}_i = Q(\bm{r}_i)$ and $\bm{r}_{i+1} = \bm{r}_i - \bm{q}_i$, until the $\mathit{L}$-th layer completes the process. The final discrete representation is the sum of all quantized codes: $\bm{z}_\mathrm{q}= \sum_{i=0}^L \bm{q}_i$, which is decoded to reconstruct motion sequences $\bm{\hat{m}} = D(\bm{z}_\mathrm{q})$. 
The core advantage of this design is the suppression of cumulative quantization errors. By propagating residuals across layers, the gradient flow in the feature space naturally decomposes as $\nabla_{z_\mathrm{e}} \mathit{L}_\mathrm{rq} = \sum_{i=0}^L \nabla_{q_i} \mathit{L}_\mathrm{rq}$. Here, rq represents the residual quantization. The first quantization layer captures the backbone features, while subsequent layers refine the residuals, enabling collaborative optimization across the hierarchy. During training, all quantization layers participate in optimizing the commitment loss:
\begin{equation}\label{residual VQ-VAE}
	\mathit{L}_{\mathrm{rq}}=\sum^L_{i=1}\Vert r_i-\text{sg}[\bm{q}_i]\Vert^2_2.
\end{equation}
Furthermore, layers are randomly dropped during training with probability $p$, encouraging the model to prioritize robust shallow feature learning and to improve generalization.

\noindent\textit{Kinematic Constraint Block}
While residual quantization mitigates information loss in discrete representations, motion representations based on the SMPL format inherently involve relative rotations and skeletal tree propagation. Minor root node rotations can induce significant artifacts in foot kinematics, manifesting as unnatural gait patterns such as foot sliding. This section introduces a kinematic constraint block (KCB) that addresses the domain discrepancy between nonlinear motion spaces and linear contact detection caused by root rotation errors. By integrating kinematic modeling with cross-attention mechanisms, our approach resolves these inconsistencies.

Given the motion sequence $\bm{\hat{m}}$ output by the decoder, we adopt a two-stage process from T2M~\cite{guo2022generating}: kinematic parameter decoupling and skeletal chain propagation. This maps discrete latent representations to physically interpretable kinematic parameter spaces, generating 3D skeletal coordinates $p \in \mathbb{R}^{n \times J \times 3}$, where $J$ is the number of joints. Ankle and foot coordinates undergo thresholding to produce binary contact labels. Through differential computation of the skeletal chain, we derive joint velocities and dynamically adjust contact confidence via sigmoid modulation. By fusing contact features, velocities, and coordinates through cross-attention mechanisms, multi-scale motion constraints are constructed to rectify decoder outputs while preserving kinematic plausibility. 

To integrate kinematic constraint into the VQ-VAE optimization framework, we redesign the reconstruction loss to account for KCB adjustments as follows:
\begin{equation}\label{ refine VQ loss}
	\mathit{L}_{vq'}=\Vert \bm{m}-KCB(\bm{\hat{m}})\Vert^2_2+\Vert \text{sg}[\bm{z}_\mathrm{e}]-\bm{z}_\mathrm{q}\Vert^2_2+\gamma\Vert \bm{z}_\mathrm{e}-\text{sg}[\bm{z}_\mathrm{q}]\Vert^2_2.
\end{equation}

\noindent\textit{Loss Function}
The total loss function for the proposed TCaS-VQ-VAE architecture integrates three components: (1) The refined VQ-VAE loss that ensures high-fidelity reconstruction, aligns encoder outputs with discrete codebook entries, and forces encoder outputs to remain close to their nearest codebook vectors. (2) The cycle-consistency loss that ensures closed-loop temporal alignment through classification-based mapping verification and refines temporal precision via regression-based error correction. (3) The residual quantization loss that suppresses discretization artifacts through multi-stage residual approximation. The final objective combines these components through weighted summation:
\begin{equation}\label{TCaS_loss}
	\mathit{L}_{\mathrm{TCaS}}=\mathit{L}_\mathrm{vq'}+\alpha_{t}\mathit{L}_{\mathrm{tcc}}+\beta_r\mathit{L}_\mathrm{rq},
\end{equation}
where $\alpha_t$ and $\beta_r$ denote the weighting coefficients for temporal consistency enforcement and residual error mitigation, respectively.

\subsection{Masked Motion Transformer}
\label{masked motion transformer}
This stage accomplishes progressive generation from textual semantics to physically plausible motions via a hierarchical decoupling mask generation strategy and dynamic alignment of residual quantized representations~\cite{guo2024momask}. The core architecture comprises two cascaded transformer modules: (1) the motion transformer for generating base-layer motion tokens and (2) the residual transformer that predicts residual-layer tokens hierarchically. Both modules engage with CLIP text features through cross-modal attention mechanisms, establishing a unified text-motion optimization framework.

The motion transformer $p_\mathrm{m}$ processes a randomly masked motion token sequence $\bm{\tilde{X}^{(0)}} = [x_i^{(0)}]_{i=1}^n \in \mathbb{R}^{n \times d}$ alongside text embeddings $\bm{w} \in \mathbb{R}^{s \times d}$. It models global motion dependencies via self-attention mechanisms while incorporating textual semantic constraints through cross-modal attention. Text embeddings are linearly projected into the motion feature space and concatenated with masked motion tokens to form augmented features $[\bm{w}; \bm{\tilde{X}^{(0)}}] \in \mathbb{R}^{(s + n) \times d}$, which are then fed into the transformer encoder. The encoder's output undergoes a softmax operation to produce probability distributions over base-layer motion tokens. The base-layer motion token prediction loss $\mathit{L}_\mathrm{mt}$, which quantifies the generation quality of motion backbone, is defined as
% \begin{equation}\label{motion transformer}
	% 	\mathcal{L}_{mt}=-\sum^n_{i=1}\log p_m(x^{(0)}_i|w, \widetilde{X}^{(0)}).
	% \end{equation}
\begin{equation}\label{motion transformer}
	\mathit{L}_\mathrm{mt}=-\sum^n_{i=1}\log p_m\!(x^{(0)}_i \mid \bm{w}, \bm{\widetilde{X}^{(0)}}).
\end{equation}
This loss optimizes the mapping between textual semantics and motion patterns by maximizing the log-probabilities of correct motion tokens.

The residual transformer $p_\mathrm{r}$ processes the motion token sequence predicted by preceding layers to hierarchically predict residual tokens. Each residual layer corresponds to a discretization level within the residual quantization module (RQM), where token granularity progressively refines from coarse to fine. For the $l$-th layer, the input comprises the accumulated predictions from previous $l-1$ layers combined with text embeddings $w$, generating the current-layer residual $\Delta x^{(j)}$. Through residual connections, predictions are aggregated across layers to form the complete motion sequence $\bm{X^{(l)}} = \bm{X^{(l-1)}} + \Delta x^{(j)}$, which serves as input to the subsequent layer $\bm{\tilde{X}^{(l+1)}} = \bm{X^{(l)}}$. The residual layer prediction loss $ \mathit{L}_\mathrm{rt}$ enforces fine-grained optimization of hierarchical tokens and is defined as
% \begin{equation}\label{residual transformer}
	% 	\mathcal{L}_{rt}=-\sum^L_{j=1}\sum^n_{i=1}\log p_r(x^{(j)}_i|w, \widetilde{X}^{(j)},j).
	% \end{equation}
\begin{equation}\label{residual transformer}
	\mathit{L}_\mathrm{rt}=-\sum_{j=1}^L \sum_{i=1}^n \log p_r\!(x^{(j)}_i \mid \bm{w}, \bm{\widetilde{X}^{(j)}}, j).
\end{equation}
% \includegraphics[width=1.0\linewidth]{inference.png}
% \caption{\textbf{Inference Process.} (a) The Motion Transformer iteratively predicts high-confidence action tokens layer by layer based on confidence scores from fully masked inputs. (b) The Residual Transformer starts predicting residual tokens from the bottom layer upward. (c) The decoder performs feature fusion of the primary action tokens and multi-layer residual tokens to generate coherent and precise motion sequences. See Section \ref{sec:method}.}
% \label{fig:inference}

% As illustrated in the Figure~\ref{fig:inference}, 
The inference pipeline commences with text embedding extraction $\bm{w} \in \mathbb{R}^{s \times d}$ via the CLIP encoder from input text. The motion sequence is initialized as a fully masked token sequence $\bm{\tilde{X}^{(0)}} = [[ \text{MASK} ]]^n_{i=1}$. The motion transformer iteratively reconstructs base-layer motion tokens $\bm{X^{(0)}}$ over $T$ iterations. In each iteration, the model predicts the probability distribution of motion tokens at the masked positions, samples candidate tokens, retains high-confidence tokens to reinforce text semantic constraints, and dynamically adjusts the mask rate based on the iteration count. Subsequently, the residual transformer hierarchically predicts residual tokens through $L$ layers, refining motion details via residual accumulation. The complete motion sequence is obtained by aggregating predictions, $\bm{X} = \bm{X^{(0)}} + \sum_{j=1}^L \Delta x^{(j)}$, which is decoded into 3D motion $\bm{\hat{m}} = D(\bm{X})$ via the TCaS-VQ-VAE decoder. Finally, the KCB polishes the generated sequence to achieve high-fidelity motion synthesis.

\section{Experiments}
\label{experiments}

This section provides comprehensive experimental validation and analysis of our TCA-T2M framework. Section~\ref{sec:exp_setups} shows the experimental setups. Section~\ref{comparison} conducts quantitative benchmarking against state-of-the-art approaches using standardized evaluation metrics across HumanML3D~\cite{guo2022generating} and KIT-ML~\cite{plappert2016kit}, demonstrating both quantitative superiority and qualitative advantages. Section~\ref{sec:ablation_study} performs systematic ablation analyses to isolate the impact of architectural components. Both experimental sections validate the technical efficacy of our framework through rigorous quantitative comparisons and qualitative evaluations.

\subsection{Datasets and Evaluation Metrics}
\label{sec:exp_setups}
We conduct experiments on two benchmark datasets for text-to-motion generation.

HumanML3D~\cite{guo2022generating} is the primary 3D motion-language dataset containing 14616 motion sequences sourced from AMASS~\cite{mahmood2019amass} and HumanAct12~\cite{guo2020action2motion} paired with 44970 text descriptions (12.1 words average). The data are standardized to 20 FPS with 10-second duration clips. Each motion sequence is annotated with at least three diverse textual scripts covering activities such as exercise and dance. 

KIT-ML~\cite{plappert2016kit} includes 3911 motion sequences from KIT~\cite{plappert2016kit} and CMU~\cite{yan2023cross}, accompanied by 6278 descriptions (8.5 words average). The data are downsampled to 12.5 FPS and segmented into 1-4 description clips. Both datasets follow the standard 80\%/5\%/15\% train/validation/test splits with standard data augmentation protocols.

We evaluate the motion generation performance using five key metrics from Ref.~\cite{guo2022generating}.
Frechet inception distance (FID) assesses global motion quality by comparing the distributional similarity between generated and real-motion features. 
R-Precision measures text-motion alignment by matching generated motions to ground-truth descriptions and reporting the Top-1/2/3 accuracies. 
Multimodal Distance (MM-Dist) quantifies cross-modal coherence as the average cosine similarity between motion features and text embeddings. MultiModality (MModality) evaluates intra-modal diversity by computing the standard deviation of 10 motion samples per text prompt, while Diversity measures global diversity through pairwise Euclidean distances between 300 randomly sampled motion pairs.
% \textbf{Frechet Inception Distance (FID)} measures distributional similarity between generated and real-motion features to assess global motion quality. \textbf{R-Precision} evaluates text-motion alignment by matching generated motions to ground-truth descriptions, reporting Top-1/2/3 accuracies for semantic fidelity quantification. \textbf{Multimodal Distance (MM-Dist)} quantifies cross-modal coherence via average cosine similarities between motion features and text embeddings. \textbf{Multimodality} assesses intra-modal diversity using standard deviation of 10 motion samples per text prompt, while \textbf{Diversity} measures pairwise Euclidean distances between 300 randomly sampled motion pairs for global diversity evaluation.

\subsection{Implementation Details}
% We employ reblocks for both the encoder and decoder, with a downscale factor of 4. 
Our models are implemented in PyTorch. For the temporal consistency-aware spatial VQ-VAE, we divide the HumanML3D dataset into 110 categories and the KIT-ML dataset into 52 categories, and set the weight of the cycle-consistency loss to 0.1. The residual VQ module consists of six quantization layers, each equipped with a codebook of 512 codes with 512-dimensional embeddings. The quantization dropout ratio is set to 0.2. Both the motion transformer and the residual transformer contain six transformer layers with six heads and a latent dimension of 384, and are applied to the HumanML3D and KIT-ML datasets. The mini-batch size is set to 256 for training the RVQ-VAE, and to 64 and 32 for training the transformers on HumanML3D and KIT-ML, respectively.
% The learning rate is linearly warmed up to 2e-4 over the first 2,000 iterations for all model trainings. 
% During inference, we use the CFG scale of 4 and 5 for M-Transformer and R-Transformer on HumanML3D, and (2, 5) on KIT-ML.

\subsection{Comparison to State-of-the-art Methods}
\label{comparison}

% \paragraph{Quantitative Results}
We compare our approach against state-of-the-art baselines~\cite{guo2022tm2t, guo2022generating, tevet2022human, zhang2023generating, chen2023executing, zhang2024motiondiffuse, pinyoanuntapong2024mmm, guo2024momask, dai2024motionlcm, zhang2024motion, huang2024controllable} on both HumanML3D~\cite{guo2022generating} and KIT-ML~\cite{plappert2016kit} datasets. Results reported in Table~\ref{tab:sota_humanmld3d} and Table~\ref{tab:sota_kit-ml} demonstrate that our method outperforms existing approaches across key fidelity metrics (FID, R-Precision, MM-Dist) while maintaining comparable performance in diversity-related measures (Diversity, MModality). The results on the two popular dataset confirm the model's capacity to generate high-fidelity motions without compromising stylistic diversity. 

In Table~\ref{tab:vae}, we further benchmark our TCaS-VQ-VAE against prior VQ-based methods~\cite{guo2022tm2t, kong2023priority, zhang2023generating, guo2024momask} on both datasets. The results establish clear superiority of our framework in both motion reconstruction and generation tasks. 

\begin{table*}[t]
	\caption{Comparison of text-to-motion generation performance on the HumanML3D dataset. The arrows $\uparrow$, $\downarrow$, and $\rightarrow$ indicate that higher, lower, and closer-to-real-motion values are better, respectively. Bold and underline indicate the best and the second best result. The model marked with * represents results obtained by reproducing the code experiments from the original paper. FID assesses global motion quality. R-Precision measures text-motion alignment. MM-Dist quantifies cross-modal coherence between motion features and text embeddings. MModality evaluates intra-modal diversity, while Diversity measures global diversity.}
	\label{tab:sota_humanmld3d}
	\centering
	\setlength{\tabcolsep}{3pt} % 
	\resizebox{\textwidth}{!}{%
		\begin{tabular}{l c c c c c c c c}
			\toprule
			\multirow{2}{*}{Method} & \multirow{2}{*}{Year} & \multicolumn{3}{c}{R-Precision $\uparrow$} & \multirow{2}{*}{FID $\downarrow$} & \multirow{2}{*}{MM Dist $\downarrow$} & \multirow{2}{*}{Diversity $\rightarrow$} & \multirow{2}{*}{MModality $\uparrow$} \\
			\cmidrule(lr){3-5}
			~ &  & Top 1 & Top 2 & Top 3 & & & & \\
			\midrule
			Real & -- & 0.511$^{\pm .003}$ & 0.703$^{\pm .003}$ & 0.797$^{\pm .002}$ & 0.002$^{\pm .000}$ & 2.974$^{\pm .008}$ & 9.503$^{\pm .065}$ & - \\ 
			\midrule
			T2M-GPT~\cite{zhang2023generating}  & 2023 & 0.492$^{\pm .003}$ & 0.679$^{\pm .002}$ & 0.775$^{\pm .002}$ & 0.141$^{\pm .005}$ & 3.121$^{\pm .009}$ & 9.722$^{\pm .082}$ & 1.831$^{\pm .048}$\\
			MotionGPT~\cite{jiang2023motiongpt} & 2023 & 0.492$^{\pm.003}$  & 0.681$^{\pm.003}$  & 0.778$^{\pm.002}$  & 0.232$^{\pm.008}$  & 3.096$^{\pm.008}$  & 9.528$^{\pm.071}$  & 2.008$^{\pm.084}$ \\
			MotionGPT~\cite{zhang2024motiongpt} & 2024 & 0.364$^{\pm.005}$ & 0.533$^{\pm.003}$ & 0.629$^{\pm.004}$ & 0.805$^{\pm.002}$ & 3.914$^{\pm.013}$ & 9.972$^{\pm.026}$ & 2.473$^{\pm.041}$ \\
			\midrule
			TM2T~\cite{guo2022tm2t}  & 2022 & 0.424$^{\pm .003}$ & 0.618$^{\pm .003}$ & 0.729$^{\pm .002}$ & 1.501$^{\pm .017}$ & 3.467$^{\pm .011}$ & \underline{9.920$^{\pm .083}$} & \underline{2.424$^{\pm .093}$} \\
			T2M~\cite{guo2022generating}  & 2022 & 0.455$^{\pm .002}$ & 0.636$^{\pm .003}$ & 0.736$^{\pm .003}$ & 1.087$^{\pm .002}$ & 3.347$^{\pm .008}$ & 9.175$^{\pm .002}$ & 2.219$^{\pm .074}$ \\
			MDM~\cite{tevet2022human}  & 2023 & 0.320$^{\pm .005}$ & 0.498$^{\pm .004}$ & 0.611$^{\pm .007}$ & 0.544$^{\pm .044}$ & 5.566$^{\pm .027}$ & 9.559$^{\pm .086}$ & \textbf{2.799$^{\pm .072}$} \\
			MLD~\cite{chen2023executing} & 2023 & 0.481$^{\pm .003}$ & 0.673$^{\pm .003}$ & 0.772$^{\pm .002}$ & 0.473$^{\pm .013}$ & 3.196$^{\pm .010}$ & 9.724$^{\pm .082}$ & 2.413$^{\pm .079}$ \\
			Mo.Diffuse~\cite{zhang2024motiondiffuse} & 2024 & 0.491$^{\pm .001}$ & 0.681$^{\pm .001}$ & 0.775$^{\pm .001}$ & 0.630$^{\pm .001}$ & 3.113$^{\pm .001}$ & 9.410$^{\pm .049}$ & 1.553$^{\pm .042}$\\
			MMM~\cite{pinyoanuntapong2024mmm}  & 2024 & 0.504$^{\pm .003}$ & 0.696$^{\pm .003}$ & 0.794$^{\pm .002}$ & \underline{0.080$^{\pm .003}$} & \underline{2.998$^{\pm .007}$} & 9.411$^{\pm .058}$ & 1.164$^{\pm .041}$ \\
			Momask$^*$~\cite{guo2024momask} & 2024  & \underline{0.508$^{\pm .003}$} & \underline{0.701$^{\pm .003}$} & 0.797$^{\pm .002}$ & 0.103$^{\pm .003}$ & 3.024$^{\pm .008}$ & 9.428$^{\pm .070}$ & 1.308$^{\pm .058}$ \\
			MotionLCM~\cite{dai2024motionlcm}  & 2024 & 0.502$^{\pm .003}$ & 0.698$^{\pm .002}$ & \underline{0.798$^{\pm .002}$} & 0.304$^{\pm .012}$ & 3.012$^{\pm .007}$ & 9.607$^{\pm .066}$ & 2.259$^{\pm .066}$\\
			Mo.Mamba~\cite{zhang2024motion} & 2024 & 0.502$^{\pm .003}$ & 0.693$^{\pm .002}$ & 0.792$^{\pm .002}$ & 0.281$^{\pm .011}$ & 3.060$^{\pm .058}$  & 9.871$^{\pm .084}$ & 2.294$^{\pm .058}$ \\
			CoMo~\cite{huang2024controllable}  & 2024 & 0.502$^{\pm .002}$ & 0.692$^{\pm .007}$ & 0.790$^{\pm .002}$ & 0.262$^{\pm .004}$ & 3.032$^{\pm .015}$ & \textbf{9.936$^{\pm .066}$} & 1.013$^{\pm .046}$ \\
			Light-T2M~\cite{zeng2025light} & 2025 & 0.511$^{\pm.003}$ & 0.699$^{\pm.002}$ & 0.795$^{\pm.002}$ & \textbf{0.040$^{\pm.002}$} & 3.002$^{\pm.008}$ & -- & 1.670$^{\pm.061}$ \\
			MG-MotionLLM~\cite{wu2025mg} & 2025 & 0.516$^{\pm.002}$ & 0.706$^{\pm.002}$ & 0.802$^{\pm.003}$ & 0.303$^{\pm.010}$ & 2.952$^{\pm.009}$ & 9.960$^{\pm.073}$ & 2.125$^{\pm.159}$ \\
			GenM$^3$~\cite{Shi_2025_ICCV} & 2025 & 0.511$^{\pm.003}$ & 0.705$^{\pm.002}$ & 0.804$^{\pm.002}$ & 0.046$^{\pm.002}$ & 2.852$^{\pm.009}$ & 9.675$^{\pm.087}$ & - \\
			\midrule
			Ours & -- & \textbf{0.517$^{\pm .003}$} & \textbf{0.709$^{\pm .002}$} & \textbf{0.806$^{\pm .002}$} & 0.068$^{\pm .003}$ & \textbf{2.951$^{\pm .007}$} & 9.485$^{\pm .090}$ & 1.103$^{\pm .048}$\\
			\bottomrule
		\end{tabular}
	}% end of resizebox
\end{table*}

\begin{table*}[t]
	\caption{Comparison of text-to-motion generation performance on the KIT-ML dataset.}
	\label{tab:sota_kit-ml}
	\centering
	\setlength{\tabcolsep}{3pt} % 
	\resizebox{\textwidth}{!}{%
		\begin{tabular}{l c c c c c c c c}
			\toprule
			\multirow{2}{*}{Method} & \multirow{2}{*}{Year} & \multicolumn{3}{c}{R-Precision $\uparrow$} & \multirow{2}{*}{FID $\downarrow$} & \multirow{2}{*}{MM Dist $\downarrow$} & \multirow{2}{*}{Diversity $\rightarrow$} & \multirow{2}{*}{MModality $\uparrow$} \\
			\cmidrule(lr){3-5}
			~ &  & Top 1 & Top 2 & Top 3 & & & & \\
			\midrule
			Real & -- & 0.424$^{\pm .005}$ & 0.649$^{\pm .006}$ & 0.779$^{\pm .006}$ & 0.031$^{\pm .004}$ & 2.788$^{\pm .012}$ & 11.08$^{\pm .097}$ & - \\ 
			\midrule
			T2M-GPT~\cite{zhang2023generating} & 2023  & 0.402$^{\pm .006}$ & 0.619$^{\pm .005}$ & 0.737$^{\pm .006}$ & 0.717$^{\pm .041}$ & 3.053$^{\pm .026}$ & 10.86$^{\pm .094}$ & 1.912$^{\pm .036}$\\
			MotionGPT~\cite{jiang2023motiongpt} & 2023 & 0.366$^{\pm.005}$ & 0.558$^{\pm.004}$ & 0.680$^{\pm.005}$ & 0.510$^{\pm.016}$ & 3.527$^{\pm.021}$ & 10.350$^{\pm.084}$ & 2.328$^{\pm.117}$ \\
			MotionGPT~\cite{zhang2024motiongpt} & 2024 & 0.340$^{\pm.002}$ & 0.570$^{\pm.003}$ & 0.660$^{\pm.004}$ & 0.868$^{\pm.032}$ & 3.721$^{\pm.018}$ & 9.972$^{\pm.026}$ & 2.296$^{\pm.022}$ \\
			\midrule
			TM2T~\cite{guo2022tm2t}  & 2022 & 0.280$^{\pm .005}$ & 0.463$^{\pm .006}$ & 0.587$^{\pm .005}$ & 3.599$^{\pm .153}$ & 4.591$^{\pm .026}$ & 9.473$^{\pm .117}$ & \textbf{3.292$^{\pm .081}$} \\
			T2M~\cite{guo2022generating} & 2022  & 0.361$^{\pm .006}$ & 0.559$^{\pm .007}$ & 0.681$^{\pm .007}$ & 3.022$^{\pm .107}$ & 3.488$^{\pm .028}$ & 10.72$^{\pm .145}$ & 2.052$^{\pm .107}$ \\
			MDM~\cite{tevet2022human}  & 2023 & 0.164$^{\pm .004}$ & 0.291$^{\pm .004}$ & 0.396$^{\pm .004}$ & 0.497$^{\pm .021}$ & 9.191$^{\pm .022}$ & 10.85$^{\pm .109}$ & 1.907$^{\pm .214}$ \\
			MLD~\cite{chen2023executing} & 2023 & 0.390$^{\pm .008}$ & 0.609$^{\pm .008}$ & 0.734$^{\pm .007}$ & 0.404$^{\pm .027}$ & 3.204$^{\pm .027}$ & 10.80$^{\pm .117}$ & \underline{2.192$^{\pm .071}$} \\
			Mo.Diffuse~\cite{zhang2024motiondiffuse} & 2024 & 0.417$^{\pm .004}$ & 0.621$^{\pm .004}$ & 0.739$^{\pm .004}$ & 1.954$^{\pm .064}$ & 2.958$^{\pm .005}$ & \textbf{11.10$^{\pm .143}$} & 0.730$^{\pm .013}$\\
			MMM~\cite{pinyoanuntapong2024mmm} & 2024  & 0.404$^{\pm .005}$ & 0.621$^{\pm .005}$ & 0.744$^{\pm .004}$ & 0.316$^{\pm .028}$ & 2.977$^{\pm .019}$ & 10.91$^{\pm .101}$ & 1.232$^{\pm .039}$ \\
			Momask$^*$~\cite{guo2024momask} & 2024  & 0.415$^{\pm .005}$ & 0.635$^{\pm .006}$ & 0.762$^{\pm .004}$ & \underline{0.228$^{\pm .009}$} & 2.889$^{\pm .020}$ & 10.612$^{\pm .088}$ & 1.208$^{\pm .042}$ \\
			Mo.Mamba~\cite{zhang2024motion} & 2024 & 0.419$^{\pm .006}$ & \underline{0.645$^{\pm .005}$} & \underline{0.765$^{\pm .006}$} & 0.307$^{\pm .041}$ & 3.021$^{\pm .025}$  & \underline{11.02$^{\pm .098}$} & 1.678$^{\pm .064}$ \\
			CoMo~\cite{huang2024controllable} & 2024 & \underline{0.422$^{\pm .009}$} & 0.638$^{\pm .007}$ & \underline{0.765$^{\pm .011}$} & 0.332$^{\pm .045}$ & \underline{2.873$^{\pm .021}$} & 10.95$^{\pm .196}$ & 1.249$^{\pm .008}$ \\
			% \RED{MG-MotionLLM~\cite{wu2025mg}} & \RED{2025} & 
			\midrule
			Ours & -- & \textbf{0.428$^{\pm .006}$} & \textbf{0.655$^{\pm .005}$} & \textbf{0.780$^{\pm .004}$} & \textbf{0.198$^{\pm .011}$} & \textbf{2.760$^{\pm .016}$} & 10.83$^{\pm .077}$ & 1.123$^{\pm .056}$\\
			\bottomrule
		\end{tabular}
	}% end of resizebox
\end{table*}

\begin{figure}[t]
	\centering
	\includegraphics[width=1.0\linewidth]{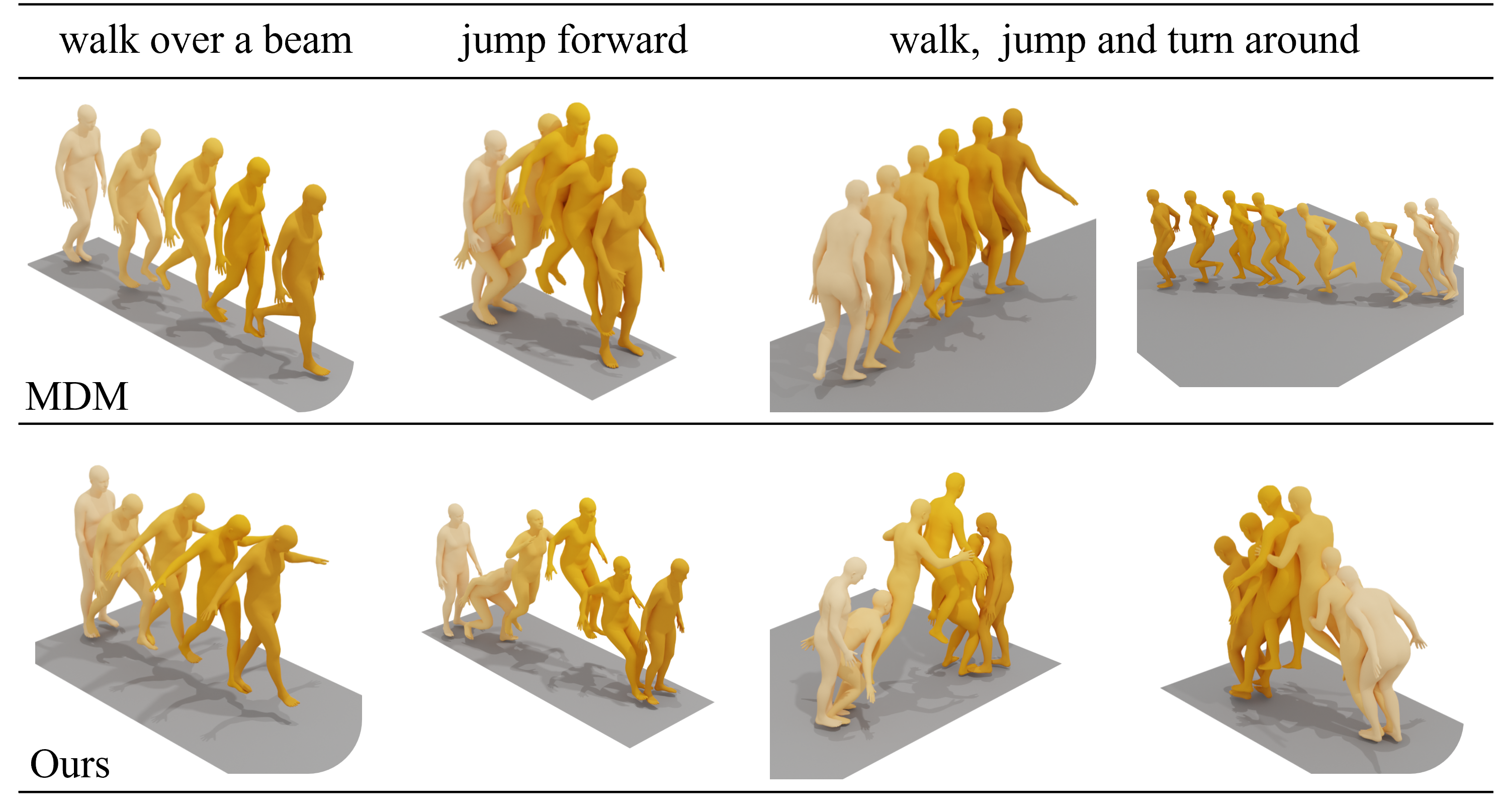}
	\caption{Qualitative comparisons between MDM\cite{tevet2022human} and our method across representative motion from the HumanML3D dataset. Key frames highlight critical motion details. The visual comparisons underscore our method's strength in semantic comprehension of textual prompts and consistent action execution across multi-step sequences with dynamic environment adaptation.}
	\label{fig:vis}
\end{figure}

% \paragraph{Qualitative Results}
Figure~\ref{fig:vis} presents qualitative comparisons between our framework and MDM~\cite{tevet2022human}. The baseline method often produces biomechanically implausible motions in complex scenarios. For example, when tasked with narrow beam traversal, MDM generates routine walking motions akin to flat-ground locomotion, failing to execute necessary balance adaptations such as arm-swaying for postural stabilization. In contrast, our model demonstrates active environmental adaptation, i.e., it autonomously detects narrow surfaces and activates compensatory postures to maintain stability. Moreover, while MDM~\cite{tevet2022human} achieves semantically plausible motion generation in simple cases (e.g., producing a complete ``crouch-takeoff-landing'' sequence for isolated ``forward jumps''), its performance degrades significantly in more complex contexts. For example, when jump actions are embedded in sequential behaviors (e.g., ``walk\textrightarrow{}jump\textrightarrow{}turn''), the generated jumps often degenerate into walking or running postures, resulting in inter-frame inconsistency. In contrast, our framework maintains core motion fidelity across both isolated and multi-step action sequences, ensuring biomechanical plausibility through dynamic environmental adaptation.
% our framework preserves core motion fidelity across multi-step action sequences—whether executed independently or integrated into complex behavior chains—ensuring biomechanical plausibility through dynamic environment adaptation.

% \setlength\tabcolsep{7.5pt}
% \renewcommand\arraystretch{1.0}
\begin{table}[t]
	\caption{Comparison of motion reconstruction and generation performance. We present an analysis of the temporal cycle-consistency constraint (TCC), the kinematic constraint block (KCB) and the residual quantization(RQ) on the HumanML3D and KIT-ML datasets.}
	\label{tab:vae}
	\centering
	\resizebox{0.5\textwidth}{!}{
		\begin{tabular}{l c c c}
			\toprule
			\multirow{2}{*}{Method} & \multicolumn{1}{c}{Reconstruction} & \multicolumn{2}{c}{Generation}  \\
			\cmidrule(lr){2-2}
			\cmidrule(lr){3-4}
			& FID$\downarrow$ & FID$\downarrow$ & MM-Dist$\downarrow$ \\
			\midrule
			\multicolumn{4}{c}{Evaluation on the KIT-ML dataset}\\
			\midrule
			M2DM\cite{kong2023priority} & 0.413$^{\pm .009}$ & 0.515$^{\pm .029}$ & 3.015$^{\pm .017}$ \\
			T2M-GPT\cite{zhang2023generating} & 0.472$^{\pm .011}$ & 0.514$^{\pm .029}$ & 3.007$^{\pm .023}$ \\
			Momask\cite{guo2024momask} & 0.184$^{\pm .002}$ & 0.228$^{\pm .009}$ & 3.021$^{\pm .025}$ \\
			Ours & \textbf{0.137$^{\pm .003}$} & \textbf{0.198$^{\pm .011}$} & \textbf{2.760$^{\pm .016}$} \\
			\midrule
			$\quad w/o$ KCB & 0.142$^{\pm .002}$ & 0.284$^{\pm .014}$ & 2.922$^{\pm .021}$ \\
			$\quad w/o$ TCC & 0.170$^{\pm .002}$ & 0.292$^{\pm .016}$ & 2.943$^{\pm .025}$ \\
			$\quad w/o$ KCB\&TCC & 0.189$^{\pm .003}$ & 0.297$^{\pm .016}$ & 2.947$^{\pm .026}$ \\
			$\quad w/o$ RQ & 0.176$^{\pm .002}$ & 0.296$^{\pm .016}$ & 2.921$^{\pm .020}$ \\
			$\quad w/o$ RQ\&KCB\&TCC & 0.198$^{\pm .003}$ & 0.309$^{\pm .017}$ & 2.958$^{\pm .027}$ \\
			\midrule
			\multicolumn{4}{c}{Evaluation on the HumanML3D dataset}\\
			\midrule
			TM2T\cite{guo2022tm2t} & 0.307$^{\pm .002}$ & 1.501$^{\pm .017}$ & 3.467$^{\pm .011}$ \\
			M2DM\cite{kong2023priority} & 0.063$^{\pm .001}$ & 0.352$^{\pm .005}$ & 3.116$^{\pm .008}$ \\
			T2M-GPT\cite{zhang2023generating} & 0.070$^{\pm .001}$ & 0.141$^{\pm .005}$ & 3.121$^{\pm .009}$ \\
			Momask\cite{guo2024momask} & 0.034$^{\pm .001}$ & 0.103$^{\pm .003}$ & 3.024$^{\pm .008}$ \\
			Ours & \textbf{0.025$^{\pm .000}$} & \textbf{0.068$^{\pm .003}$} & \textbf{2.951$^{\pm .007}$} \\
			\midrule
			$\quad w/o$ KCB & 0.042$^{\pm .000}$ & 0.105$^{\pm .004}$ & 3.010$^{\pm .010}$ \\
			$\quad w/o$ TCC & 0.054$^{\pm .000}$ & 0.111$^{\pm .004}$ & 2.994$^{\pm .009}$ \\
			$\quad w/o$ KCB\&TCC & 0.058$^{\pm .000}$ & 0.119$^{\pm .005}$ & 2.988$^{\pm .008}$ \\
			$\quad w/o$ RQ & 0.056$^{\pm .000}$ & 0.114$^{\pm .005}$ & 2.922$^{\pm .021}$ \\
			$\quad w/o$ RQ\&KCB\&TCC & 0.071$^{\pm .001}$ & 0.128$^{\pm .006}$ & 3.032$^{\pm .008}$ \\
			\bottomrule
		\end{tabular}
	}
\end{table}

% \subsection{Long Sequence Generation and Zero-shot Generation}
\subsection{Generalization Demonstration}
\label{sec:LSG&ZSG}
\textcolor{black}{To evaluate our model's generalization ability, we conduct experiments on long-sequence motion generation and zero-shot motion generation.
}

\begin{figure}[t]
	\centering
	\includegraphics[width=1.0\linewidth]{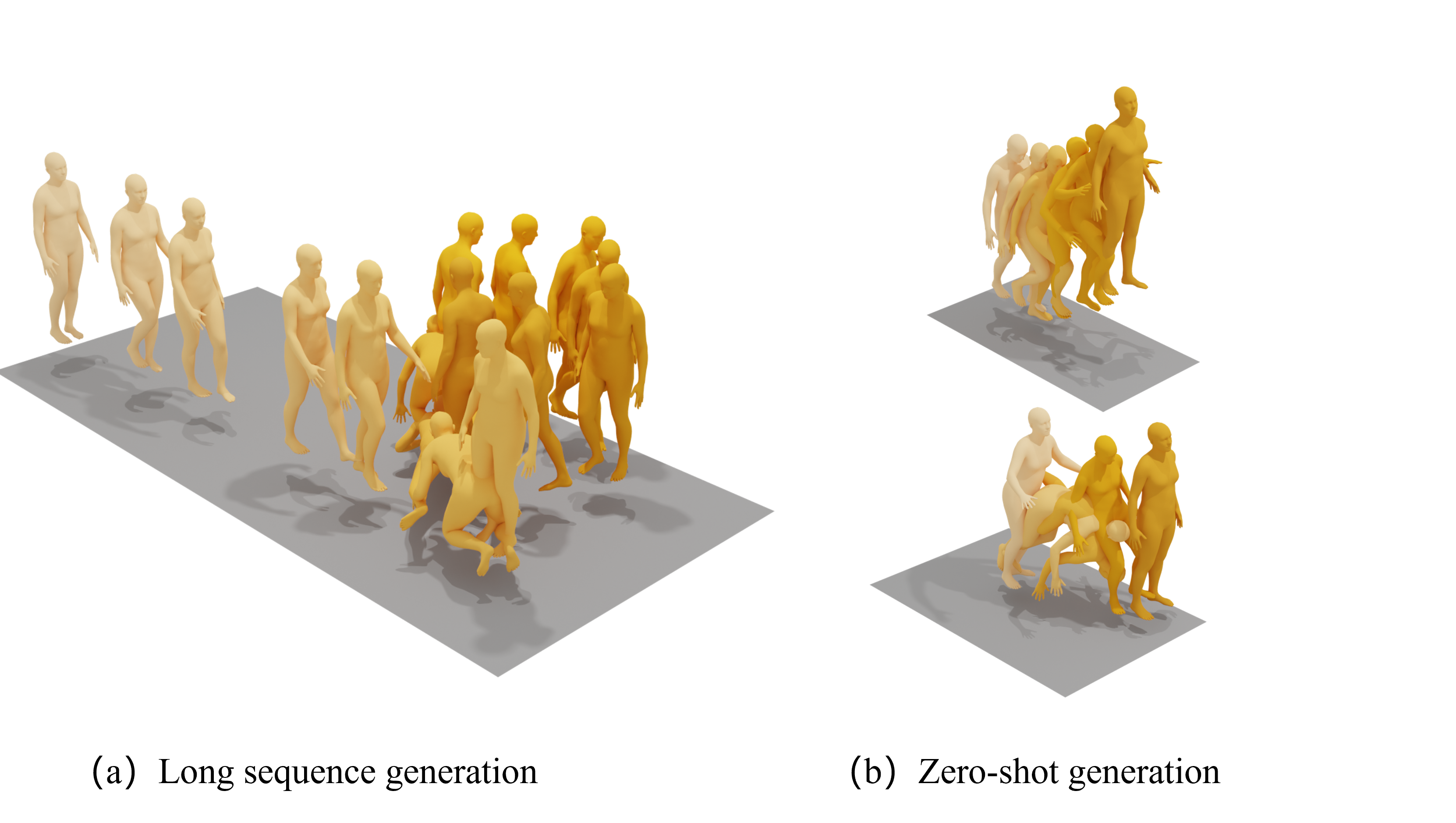}
	\caption{Visualizations of long motion generation and zero-shot motion generation. (a) Long motion generation. We integrate three text prompts---``a person walks forward then turns right'',``a person crawling from right to left'' and``the person is walking in a counterclockwise circl''—to generate a long-sequence motion. (b) Zero-shot generation. We separately test the text prompts``A person is climbing a ladde'' and``A person is rolling'' to generate zero-shot motion sequences.}
	\label{fig:gene}
\end{figure}

%\paragraph{Long Motion Generation}
Due to the short duration of motion data in the HumanML3D and KIT datasets, where most samples are under 10 seconds, creating extended motion sequences is challenging. we leverage the pre-trained masked motion model as a prior for synthesizing long motion sequences without further training. 
Long human motion generation is performed using multiple textual prompts. Our model first produces motion token sequences of arbitrary length for each prompt. Subsequently, our model generates transitional motion tokens based on the end of the prior sequence and the start of the following one. We visualize the long sequence motion generation in Fig.~\ref{fig:gene}(a).

%\paragraph{Zero-shot Motion Generation}
Zero-shot motion generation aims to generate plausible and semantically aligned human motions for previously unseen actions or motions in the training set. This task requires robust generalization and language comprehension capabilities in the model. We select samples involving actions not present in the dataset but with similar motions for zero-shot motion generation during testing. The visualization results are shown in Fig.~\ref{fig:gene}(b).
% Zero-shot motion generation requires robust generalization and language comprehension capabilities in the model. We select samples involving actions not present in the dataset but with similar motions for zero-shot motion generation testing. The visualization results are shown in Figure~\ref{fig:gene}(b).

\subsection{Ablation and Parameter Analysis}
\label{sec:ablation_study}
To validate the efficacy of our architectural innovations, we conduct systematic ablation studies on two key designs: the temporal cycle-consistency constraint (TCC) and the kinematic constraint block (KCB). Experiments are performed on the HumanML3D~\cite{guo2022generating} dataset following the same evaluation protocols as the main benchmark. % Quantitative results are presented in Table~\ref{tab:vae}, with qualitative comparisons in Table~\ref{tab:ablation}.
% Experiments employ identical evaluation protocols to those applied in the HumanML3D~\cite{guo2022generating} benchmark, with quantitative results tabulated in Table~\ref{tab:vae} and qualitative evidence presented in Table~\ref{tab:ablation}.

%\paragraph{Component Analysis} 
Table~\ref{tab:vae} presents a systematic analysis of the impact of architectural components. The results demonstrate that the TCC and the KCB substantially improve motion reconstruction fidelity and motion generation quality compared to baseline models.

\begin{table}[t]
	\caption{Ablation study on the impact of loss type, cycle length for the temporal cycle-consistency constraint, and codebook size ($K \times d$) on HumanML3D. $K$ denotes the number of codes, and $d$ is the embedding dimension. See Section~\ref{sec:ablation_study} for details.}
	\label{tab:ablation}
	\centering
	\resizebox{0.5\textwidth}{!}{
		\begin{tabular}{l c c c c}
			\toprule
			\multirow{2}{*}{Ablations} & \multirow{2}{*}{Type} & \multicolumn{1}{c}{Reconstruction} & \multicolumn{2}{c}{Generation}  \\
			\cmidrule(lr){3-3}
			\cmidrule(lr){4-5}
			& & FID$\downarrow$ & FID$\downarrow$ & MM-Dist$\downarrow$ \\
			\midrule
			\multirow{3}{*}{TCC loss} & Classification & 0.113$^{\pm .001}$ & 0.134$^{\pm .004}$ & 3.031$^{\pm .008}$ \\
			& \textbf{Regression(MSE)} & \textbf{0.025$^{\pm .000}$} & \textbf{0.068$^{\pm .003}$} & \textbf{2.951$^{\pm .007}$} \\
			& Regression(Huber) & 0.040$^{\pm .000}$ & 0.082$^{\pm .003}$ & 2.932$^{\pm .007}$ \\
			\midrule
			\multirow{4}{*}{Cycle length} & \textbf{2} & \textbf{0.025$^{\pm .000}$} & \textbf{0.068$^{\pm .003}$} & \textbf{2.951$^{\pm .007}$} \\
			& 3 & 0.030$^{\pm .000}$ & 0.103$^{\pm .004}$ & 2.997$^{\pm .008}$ \\
			& 4 & 0.027$^{\pm .000}$ & 0.095$^{\pm .004}$ & 2.993$^{\pm .007}$ \\
			& 5 & 0.039$^{\pm .000}$ & 0.122$^{\pm .004}$ & 2.995$^{\pm .007}$ \\
			& 8 & 0.026$^{\pm .000}$ & 0.105$^{\pm .004}$ & 2.965$^{\pm .009}$ \\
			\midrule
			%   & \textbf{512$\times$512} & \textbf{0.025$^{\pm .000}$} & \textbf{0.068$^{\pm .003}$} & \textbf{2.951$^{\pm .007}$} \\
			% $\#$ of code $\times$  & 1024$\times$256 & 0.023$^{\pm .000}$ & 0.115$^{\pm .005}$ & 3.009$^{\pm .009}$ \\
			% code dimension  & 2048$\times$128 & 0.077$^{\pm .000}$ & 0.402$^{\pm .007}$ & 3.109$^{\pm .010}$ \\
			%   & 4096$\times$64 & 0.582$^{\pm .002}$ & 1.323$^{\pm .017}$ & 3.445$^{\pm .009}$ \\
			
			\multirow{4}{*}{Codebook size}& \textbf{512$\times$512} & \textbf{0.025$^{\pm .000}$} & \textbf{0.068$^{\pm .003}$} & \textbf{2.951$^{\pm .007}$} \\
			& 1024$\times$256 & 0.023$^{\pm .000}$ & 0.115$^{\pm .005}$ & 3.009$^{\pm .009}$ \\
			& 2048$\times$128 & 0.077$^{\pm .000}$ & 0.402$^{\pm .007}$ & 3.109$^{\pm .010}$ \\
			& 4096$\times$64 & 0.582$^{\pm .002}$ & 1.323$^{\pm .017}$ & 3.445$^{\pm .009}$ \\
			\bottomrule
		\end{tabular}
	}
\end{table}

% \paragraph{Temporal Cycle-Consistency Constrain Loss} 
As discussed in Section~\ref{sec:tcas_vq_vae}, the TCC $\mathcal{L}_{\mathrm{tcc}}$ can be realized in different ways. Here, we evaluate the impact of different loss functions for realizing $\mathcal{L}_{\mathrm{tcc}}$. Specifically, we experiment on a classification-based cycle loss and two regression-based formulations, namely, mean squared error (MSE) and huber loss (with $\delta = 0.1$). As shown in Table~\ref{tab:ablation}, MSE-based regression outperforms both alternatives across all metrics, achieving the lowest reconstruction FID (0.025) and generation FID (0.068), alongside superior multimodal distance (MM-Dist: 2.951). These results indicate that directly regressing motion features promotes more precise temporal alignment than classification-based or Huber-based regression losses, underscoring the effectiveness of MSE for enforcing cycle consistency.
% We conduct experiments on the HumanML3D~\cite{guo2022generating} dataset to compare the performance of distinct loss functions. Specifically, we implement classification cycle loss and regression cycle loss. For regression cycle loss, we further explore two optimization strategies: direct Mean Squared Error (MSE) minimization and Huber loss with $\delta =0.1$ balancing gradient sensitivity for small residuals and outlier resilience. The results demonstrate that the MSE-based regression cycle loss outperforms both the classification cycle loss and Huber-based regression cycle loss in terms of motion reconstruction fidelity and generation diversity. 

%\paragraph{Cycle Length} 
To systematically evaluate the temporal dependency modeling capabilities, we conduct ablation studies on cycle length $L$. When $L=2$, the temporal cycle follows a simple loop $[0,1,0]$, representing a closed-loop traversal from sequence 0 to 1 and back. When $L=3$, the loop extends to a multi-stage sequence $[0,1,2,0]$, introducing additional temporal context for sequential pattern learning. As shown in Table~\ref{tab:ablation}, cycle length has a minimal effect on reconstruction and generation metrics, with $L=2$ achieving the best balance (reconstruction FID: 0.025, generation FID: 0.068). However, increasing $L$ adds computational overhead without significant gains, motivating our choice of $L=2$ for efficiency and performance.
% To systematically evaluate the temporal dependency modeling capabilities, we conduct ablation studies on cycle length configurations. When cycle length is set to $2$, the resulting cycles exhibit a structure of $[0,1,0]$, representing a closed-loop traversal from sequence 0 to 1 and back. Extending the cycle length to $3$ creates a multi-stage sequence $[0,1,2,0]$, introducing additional temporal context for sequential pattern learning. Experimental results reveal that while cycle length has negligible impact on model performance metrics, increasing cycle length leads to growth in computational overhead for loss calculation. This efficiency-accuracy balance motivates our adoption of cycle length of $2$, which achieves state-of-the-art performance while maintaining computational efficiency.

\begin{table}[t]
	\caption{Evaluation of temporal consistency for human motion generation. The training time is reported as the average time per iteration on an NVIDIA RTX 4090 GPU.}
	\label{tab:temp_cons}
	\centering
	\begin{tabular}{l|cc|c}
		\toprule
		\multirow{2}{*}{Method} & \multicolumn{2}{c|}{Kendall's tau $\uparrow$} & \multirow{2}{*}{\makecell{Training \\time (s)}} \\
		\cmidrule(lr){2-3}
		& HumanML3D & KIT-ML & \\
		\midrule
		Ours w/o TCC & 0.1757 & 0.3524  & 0.26 \\
		Ours         & \textbf{0.2571} & \textbf{0.4820} & 0.06 \\
		\bottomrule
	\end{tabular}
\end{table}

%\paragraph{Codebook Size} 
We examine the trade-off between codebook capacity and embedding dimension by varying the number of codes $K$ and their corresponding dimensions $d$ while keeping total capacity approximately constant. Table~\ref{tab:ablation} shows that a $512$-entry codebook with $512$-dimensional embeddings achieves the best results across metrics (reconstruction FID: 0.025, generation FID: 0.068). While increasing the number of codes from $256$ to $1024$ or $2048$ marginally improves reconstruction fidelity, it significantly degrades generation quality, as reflected by higher FID and MM-Dist scores. Conversely, reducing embedding dimensions (e.g., $4096\times64$) leads to substantial performance degradation, likely due to over-compression and loss of fine-grained motion details. Based on these findings, we adopt the $512\times512$ configuration for high-fidelity motion generation.
% To evaluate the impact of codebook design, we investigate the interplay between codebook capacity and embedding dimension through ablation studies. Table~\ref{tab:ablation} demonstrates that increasing codebook entries within fixed capacity marginally improves motion reconstruction fidelity, but reduces generation ability. As the number of code increases, the dimension of per code decreases, leading to over-compression of feature information and consequent degradation of reconstruction and generation capabilities. Based on these results, we select a 512-entry codebook with 512-dimensional embeddings to achieve the high fidelity motion generation.

\begin{table}[t]
	\caption{Comparison of model parameters, inference efficiency, and motion generation quality.}
	\label{tab:time}
	\centering
	\begin{tabular}{l|cc|c|c}
		\toprule
		\multirow{2}{*}{Method} & \multicolumn{2}{c|}{Model size (M)} & \multirow{2}{*}{FID$\downarrow$} & \multirow{2}{*}{\makecell{Inference\\time (s)}} \\
		\cmidrule(lr){2-3}
		& Tokenizer & Total & & \\
		\midrule
		TM2T~\cite{guo2022tm2t} & 36.71 & 131.01 & 1.50 & 0.46 \\
		T2M~\cite{guo2022generating} & 3.44 & 40.29 & 1.09 & 0.02\\
		% MDM~\cite{tevet2022human} & 0.54 & 11.03\\
		T2M-GPT~\cite{zhang2023generating} & 19.44 & 32.86 & 0.12 & 0.23\\
		% MLD~\cite{chen2023executing} & 0.47 & 0.13\\
		% Mo.Diffuse~\cite{zhang2024motiondiffuse} & 0.63 & 6.60\\
		\midrule
		Ours & 44.90 & 71.37 & 0.07 & 0.09\\
		\bottomrule
	\end{tabular}
\end{table}

\begin{figure*}[t]
	\centering
	\includegraphics[width=1.0\linewidth]{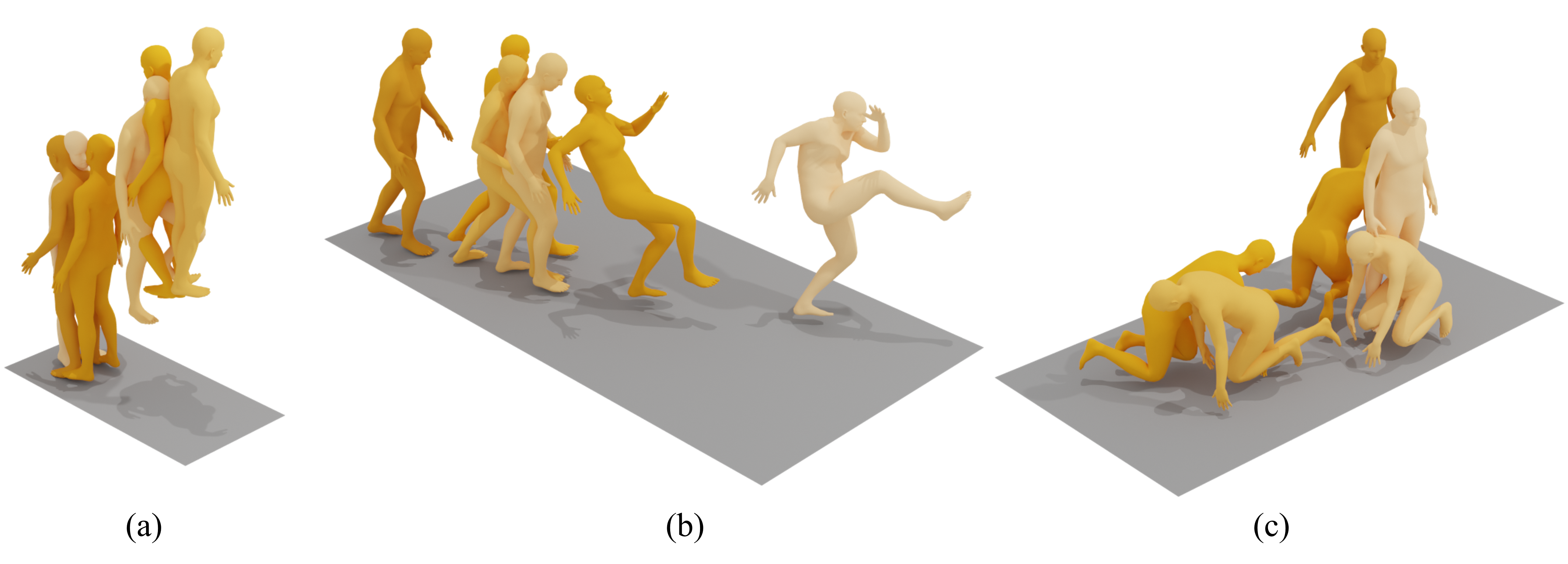}
	\caption{Visualizations of failure cases of our approach. (a) shows that a person walks up stairs. Then the person turns right and walks back down stairs. (b) shows a man walking backwards. Then, he punches and kicks. (c) shows a man getting down on his hands and feet and crawling forward. Then, he turns around and crawls back before standing up again. }
	\label{fig:failure}
\end{figure*}

%\paragraph{Evaluation of Temporal Consistency} 
To evaluate the temporal consistency of motion generation, we adopt Kendall’s tau~\cite{dwibedi2019temporal} as the evaluation metric. This metric measures the rank correlation between frames in the generated sequence and the corresponding frames in the reference sequence, and it captures how well the temporal order is preserved. Quantitative evaluations of temporal consistency are presented in Table~\ref{tab:temp_cons}. When the TCC constraint is removed, the value of this metric drops considerably, demonstrating that our method helps improve the temporal consistency of text-to-motion generation. 

%\paragraph{Comparison of Training Cost}
With the TCC constraint, the per-batch training time increases from 0.06 s to 0.26 s. Although this introduces extra training cost, the model size remains unchanged, and therefore the inference time is not affected.

%\paragraph{Comparison of Model Parameters and Inference Time} 
Analysis and comparison on parameters and inference efficiency are summarized in Table~\ref{tab:time}. The inference cost is computed as the average inference time for 100 samples on a single NVIDIA 4090 device. Compared to baseline methods, our model achieves a more favorable balance between generation quality and efficiency.
% In Table~\ref{tab:time} , we assess the efficiency and quality of motion generation across different baselines. 

%\paragraph{Failure Case Analysis} 
Figure~\ref{fig:failure} illustrates that failure cases of our text-to-motion generation approach. We select text prompts that contain multiple consecutive abrupt motions, where substantial pose changes may occur between adjacent frames. We find that due to the complexity of certain text prompts, the generated motions do not align well with the intended descriptions, and some frames exhibit implausible human poses. These examples pose significant challenges for current state-of-the-art text-to-motion generation models and highlight directions for future research.

\section{Conclusion and Limitations}
This paper presented TCA-T2M, a framework for temporal consistency-aware text-to-motion generation. By integrating cyclic temporal alignment constraints into discrete motion representation learning, TCA-T2M enables the encoder to capture invariant temporal structures across sequences of the same action. Our temporal consistency-aware spatial VQ-VAE (TCaS-VQ-VAE) ensures that corresponding motion phases align in the latent space, while the kinematic constraint block enhances the physical plausibility of generated motions by enforcing smooth joint dynamics. Extensive experiments on HumanML3D and KIT-ML demonstrate that TCA-T2M outperforms existing methods, generating temporally coherent, semantically aligned, and physically realistic motion sequences.

Although our approach improves the quality of text-to-motion generation, two key challenges persist in this field. First, semantic comprehension errors remain, with rare cases where the generated motion completely contradicts the intended meaning. Second, due to the limited availability of datasets, motion diversity is insufficient. Moreover, since most dataset motions are short, generating long sequences in real time remains a critical area for future research.

\vspace{1cm}
\noindent \textbf{Abbreviations:} \\
CLIP, contrastive language-image pre-training; Seq2Seq, sequence to sequence; VQ-VAE, vector quantized variational autoencoder

\noindent \textbf{Declarations:} \\
\begin{small}
	\noindent \textbf{Data Availability:} The datasets used in this study are publicly available at: HumanML3D: \url{https://github.com/EricGuo5513/HumanML3D}; KIT-ML: \url{https://motion-annotation.humanoids.kit.edu/dataset/}. The code is available at \url{https://github.com/Giat995/TCA-T2M/}
	
	\noindent \textbf{Competing Interests:} Xin Geng is an Associate Editors at Visual Intelligence and was not involved in the editorial review of this article or the decision to publish it. The authors declare
	that they have no other competing interests.
	
	\noindent \textbf{Author Contributions:} All the authors contributed to the study conception and design. Conceptualization of this study, methodology and analysis were performed by Hongsong Wang, Wenjing Yan and Qiuxia Lai. All the authors read and approved the final manuscript.
	
	\noindent \textbf{Funding: }This research was supported by the National Natural Science Foundation of China (Nos. 52441503 and 62302093) and 
	the Natural Science Foundation of Jiangsu Province (No. BK20230833).
	
	\noindent \textbf{Acknowledgements:} We would like to thank the Big Data Computing Center of Southeast University for their support in facilitating the numerical calculations. 
	
\end{small}

% BibTeX from reference.bib
%\bibliographystyle{sn-apacite}
\bibliographystyle{vi}
\bibliography{reference}

\end{document}